# Exploring consumers' response to text-based chatbots in e-commerce: the moderating role of task complexity and chatbot disclosure

Xusen Cheng
*Renmin University of China, Beijing, China*
Ying Bao
*University of International Business and Economics, Beijing, China*
Alex Zarifis
*Loughborough University, Loughborough, UK*
Wankun Gong
*Beijing University of Chemical Technology, Beijing, China, and*
Jian Mou
*Management Information System, Pusan National University, Kumjeong-ku, Republic of Korea*

## Abstract

**Purpose** – Artificial intelligence (AI)-based chatbots have brought unprecedented business potential. This study aims to explore consumers' trust and response to a text-based chatbot in e-commerce, involving the moderating effects of task complexity and chatbot identity disclosure.

**Design/methodology/approach** – A survey method with 299 useable responses was conducted in this research. This study adopted the ordinary least squares regression to test the hypotheses.

**Findings** – First, the consumers' perception of both the empathy and friendliness of the chatbot positively impacts their trust in it. Second, task complexity negatively moderates the relationship between friendliness and consumers' trust. Third, disclosure of the text-based chatbot negatively moderates the relationship between empathy and consumers' trust, while it positively moderates the relationship between friendliness and consumers' trust. Fourth, consumers' trust in the chatbot increases their reliance on the chatbot and decreases their resistance to the chatbot in future interactions.

**Research limitations/implications** – Adopting the stimulus–organism–response (SOR) framework, this study provides important insights on consumers' perception and response to the text-based chatbot. The findings of this research also make suggestions that can increase consumers' positive responses to text-based chatbots.

**Originality/value** – Extant studies have investigated the effects of automated bots' attributes on consumers' perceptions. However, the boundary conditions of these effects are largely ignored. This research is one of the first attempts to provide a deep understanding of consumers' responses to a chatbot.

**Paper type** Research paper

This work is supported by the Fundamental Research Funds for the Central Universities, and the Research Funds of Renmin University of China (Grant No.21XNO002).

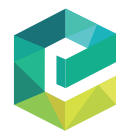

## 1. Introduction

Disembodied conversational agents (DCAs) have now become quite common (Araujo, 2018; Luo *et al.*, 2019). Galvanized by artificial intelligence (AI) and machine learning, DCAs in the form of chatbots can simulate human interactions through text chats or voice commands in various fields and have brought unprecedented business potential (Luo *et al.*, 2019). A chatbot is defined as "a machine conversation system which interacts with human users via natural conversational language" (Shawar and Atwell, 2005, p. 489). Specifically, the market size of chatbots is expanding from $250m in 2017 to over $1.34bn by 2024 (Pise, 2018). Many famous brands and major platforms, such as eBay, Facebook, WeChat, Amazon and Apple's Siri, have rolled out chatbots to take orders, recommend products or provide other customer services (Thompson, 2018; Go and Sundar, 2019). Gartner predicts that by 2021, almost 15% of the customer services will be completely handled by AI (Mitchell, 2018).

AI-enabled chatbots have provided several unique business benefits in the e-commerce setting. First, the chatbots are able to understand consumers' requests and automate the customer services, addressing their requests effectively (Daugherty *et al.*, 2019). Moreover, compared to a human, the AI-enabled chatbot will not experience negative emotions or work exhaustion and can always interact with consumers in a friendly manner (Luo *et al.*, 2019). Additionally, the chatbots can easily handle a large number of customer communications at the same time and increase the customer-service efficiency.

While the adoption of AI in e-commerce can bring several business benefits for the companies or platforms from the perspective of the supply side, a key challenge for the application of chatbots is the customers' perception from the perspective of the demand side (Froehlich, 2018). Although the chatbots can interact with consumers in a more friendly way compared with the customer service of employees, attributes of the technology have created customer pushback. For example, consumers may find it uncomfortable talking with a technology-based chatbot for their specific personal needs or purchasing decisions. Moreover, consumers may have a prejudice that the computer programs lack empathy and personal feelings, perceiving the chatbots as less trustworthy than humans. This negative perception of AI-enabled chatbots will even have negative effects on their perception of the brands or companies. Therefore, firms should apply the chatbot wisely to provide better customer service. For instance, firms can increase effectiveness by introducing chatbots to complete easy customer requests, but the technology might not be applied for complex tasks when customers have very demanding requests, which need employees to intervene (e.g. service recovery, price negotiation). As a result, there is a call for a deeper understanding of consumers' perceptions and responses to a text-based chatbot in the e-commerce setting, involving the effects of the chatbots' specific attributes.

Extensive research exists on AI-enabled systems in business practice, including social robots (de Kervenoael *et al.*, 2020), machine "teammates" (Seeber *et al.*, 2020) and chatbots (Luo *et al.*, 2019). Findings indicate that attributes of the AI-enabled computer programs, such as anthropomorphism (Castelo *et al.*, 2019; Desideri *et al.*, 2019; Yuan and Dennis, 2019), humanness (Wirtz *et al.*, 2018), empathy (de Kervenoael *et al.*, 2020) and adaptivity (Heerink *et al.*, 2010) will significantly impact consumers' perception, decision-making (Duan *et al.*, 2019), purchase intention (Mende *et al.*, 2019) and use intention (Heerink *et al.*, 2010; de Kervenoael *et al.*, 2020). Although existing studies have drawn attention to consumers' positive perception of the application of text-based chatbots in the e-commerce setting, current empirical evidence on consumers' pushback is still limited. Moreover, consumers' perceptions or responses may differ in different conditions. While the boundary conditions of consumers' perception and response behavior have been largely ignored in existing studies. Against this backdrop, we address both the practical and theoretical aspects with the following research questions:

*RQ1.* What is the impact of the unique attributes of a text-based chatbot on consumer responses to the text-based chatbot in the e-commerce context?

*RQ2.* How does the task complexity and chatbot identity disclosure moderate the relationship between these attributes and consumer responses?

*RQ3.* How does consumers' trust impact their responses to the text-based chatbot in e-commerce?

In order to answer these research questions, we draw upon the stimulus–organism–response (SOR) framework (Mehrabian and Russell, 1974) and test a model of consumers' trust and subsequent responses to a text-based chatbot in the e-commerce setting. We also integrate the moderating role of task complexity and consumers' perception of the disclosure of the chatbot to test the boundary conditions of different consumer responses. The SOR framework provides an ideal theoretical framework to examine the links between stimulus, organism and response systematically (Kim *et al.*, 2020). Therefore, it is suitable to address our research questions.

This paper is structured as follows. First, we introduce the theoretical background and present the hypotheses development. Next, we present the research methodology and data analysis results. Finally, we conclude with a brief summary of the findings, the theoretical contributions and the practical implications. Limitations and future research are also provided.

## 2. Theoretical background and hypotheses development
### 2.1 Stimulus–organism–response model
The "stimulus-organism-response framework" was initially proposed by Mehrabian and Russell (1974) and then modified by Jacoby (2002). The SOR framework suggests that individuals will react to the environmental factors and have certain behaviors (Kamboj *et al.*, 2018). Specifically, individuals' reaction behaviors consist of two typical forms: approach behaviors with positive actions, such as the preference to affiliate, explore and averting behaviors with the opposite actions, such as the desire to react negatively (Kim *et al.*, 2020). Existing studies have extended the SOR framework to many areas, including the online shopping experience (Eroglu *et al.*, 2003), website experience (Mollen and Wilson, 2010), consumer behavior in the retail industry (Rose *et al.*, 2012) and tourism industry (Kim *et al.*, 2020). Among the abovementioned contexts, e-commerce is one of the most representative. The SOR theory has been adopted in the e-commerce context as a psychology theory to investigate consumer behavior (Fiore and Kim, 2007; Chang *et al.*, 2011; Wu and Li, 2018). For example, Eroglu *et al.* (2003) argued that the environmental or atmospheric cues of the online platform (*S*-stimuli) can affect consumers' cognition or emotion (*O*-organism), which will then influence their behavioral outcomes (*R*-response). Similarly, in this research, we adopted the SOR framework to explain consumers' organism process and behavior outcomes toward the text-based chatbot in the e-commerce setting.

The SOR framework consists of three elements: stimulus, organism and response (Kamboj *et al.*, 2018). Similar to the "input-response-output framework", the "stimulus," "organism" and "response" correspond to the "input," "response" and "output." More specifically, "stimulus" is defined as the environmental factors encountered by individuals (Jacoby, 2002), which will arouse them (Eroglu *et al.*, 2001). In this research setting, "stimulus" is defined as the unique attributes of the text-based chatbot that will arouse the consumers. The second element "organism" can be defined as individuals' affective and cognitive condition and their process of subconsciously intervening between the stimuli and responses (Jacoby, 2002). For example, in the commercial setting, the "organism" includes the individuals' cognitive and affective perceptions of their experience as a consumer (Rose *et al.*, 2012). The third element "response" has been described as individuals' behavioral consequences or outcomes to the environment (Donovan and Rossiter, 1982). As aforementioned, individuals' response to the environment includes both positive actions and negative actions. For example, in the commercial setting, a positive response may include consumers' willingness to stay, explore or affiliate (Bitner, 1992), while a negative response may refer to individuals' willingness to react negatively. In this research setting, we take both consumers' positive and negative reactions toward the text-based chatbot into consideration.

## 2.2 Stimulus: unique attributes of a text-based chatbot

Turing proposed the Turing test to determine AI or machine intelligence in 1950 (Turing and Haugeland, 1950). A machine that passed the Turing test can be identified as more realistic and human-like, which makes it more compelling for consumers (Gilbert and Forney, 2015). Thus, the realistic and human-like characteristics of the chatbots will have a great impact on the usability of the chatbot. In the e-commerce service, AI has been applied in the form of text-based chatbots that can respond to consumers' requests automatically (Kaplan and Haenlein, 2019). The AI-enabled text-based chatbot has been widely used in the online purchasing process. On the one hand, the service provided by a text-based chatbot can be more customer-friendly when compared with some traditional human customer service. According to social response theory (Nass et al., 1994) and anthropomorphism theory (Blut et al., 2021), customers will perceive higher satisfaction during the interaction with human-like machines. While on the other hand, some consumers perceive that these chatbots always give rigid or standard answers and are not empathetic enough. As a result, we mainly focus on the effects of these two attributes (empathy and friendliness) of a text-based chatbot on consumers' outcomes.

*2.2.1 Empathy of the chatbot.* Empathy can be defined as "ability to identify, understand, and react to others' thoughts, feelings, behavior, and experiences" (Murray et al., 2019, p. 1366). It is a multidimensional construct that covers both cognitive and affective perceptions (de Kervenoael et al., 2020; Powell and Roberts, 2017). In extant research on service management and information systems, empathy can be defined as the required skills for successful interactions between consumers and robots (Birnbaum et al., 2016). According to Czaplewski (2002)'s RATER model, empathy is also categorized as one of the five dimensions (reliability, assurance, tangibles, empathy and responsiveness) of service quality (Czaplewski et al., 2002). In the traditional service setting, employees have been trained to always take the consumers' point of view. Even if many consumers may ask similar questions every day, professional employees are not supposed to be impatient and they are trained to always be empathetic and offer alternatives to address consumers' problems. In the online service setting, some text-based chatbots have been set to automatically reply to consumers when there is a need. The chatbot with a high level of empathy can understand the consumers' needs better and provide alternatives to address their issues. As a result, consumers will be more likely to have a positive perception of the services provided by the chatbot.

*2.2.2 Friendliness of the chatbot.* In a traditional business setting, employees are required to be friendly to consumers and convey positive emotions to them (Tsai and Huang, 2002). Friendly and helpful service providers can please consumers and increase their positive perception of the service quality (Chen et al., 2012). Following Pugh (2001), when employees convey more positive emotions, consumers will be more satisfied with the service, resulting in their positive response to the store, such as repurchase and positive word of mouth. In previous research on consumers' acceptance of robots, users gave high ratings on the effects of friendliness and trust (Fridin and Belokopytov, 2014). When interacting with the text-based chatbots, rigid or indifferent responses from the chatbot will result in consumer pushback. On the other hand, a friendly manner will positively impact consumers' moods and perceptions of the store and brands (Tsai and Huang, 2002).

*2.3 Organism: trust toward the text-based chatbot*
As aforementioned, "organism" refers to individuals' cognitive or affective condition. The formation of trust also involves individuals' emotions and feelings. For example, individuals may prefer not to trust others simply because of their perceived bad feelings (Hoff and Bashir, 2015). Therefore, the process of consumers' trust formation on the text-based chatbot can be regarded as the "organism" in this research setting.

Specifically, trust can be defined as the individual or group's willingness to be vulnerable to the other party (Hoy and Tschannen-Moran, 1999; Mayer *et al.*, 1995). Generally, trust has been widely discussed in many research fields in recent years, such as psychology, economics, sociology and information systems (Breuer *et al.*, 2016; Ert *et al.*, 2016). In extant research, trust has been investigated mostly in the context of interpersonal interaction. Nowadays, with the growing interaction between humans and robots, trust can also be applied to explain the way people interact with technology (Hoff and Bashir, 2015). In this research, we mainly focus on consumers' trust toward the text-based chatbot during the online purchasing process.

Extant research has investigated the antecedents of trust from several lenses. The most widely accepted antecedents of trust are ability, integrity and benevolence (Jarvenpaa *et al.*, 1997). Considering trust between humans and autonomous machines, anthropomorphic characteristics will impact users' trust perception (Złotowski *et al.*, 2017). According to social response theory (Nass *et al.*, 1994), the interactions between human and computer are fundamentally social (Adam *et al.*, 2020). Even if the consumers know that the computer does not hold feelings or emotions as humans do, they still tend to perceive the computers or machines as social actors. The theory has also provided evidence on studies investigating how humans adopt social rules to the anthropomorphically designed machines (Adam *et al.*, 2020). In other words, humans will react similarly to those machines with human characteristics as they react to humans (Ischen *et al.*, 2019). Moreover, the anthropomorphism theory also argues that individuals' human-like perceptions of the robots will facilitate the human–robot interaction, thus increasing customers' positive perception of the interactions (Blut *et al.*, 2021). As aforementioned, we mainly focus on empathy and friendliness as the anthropomorphic attributes of the chatbots. Based on social response theory and anthropomorphism theory, we can infer that consumers may perceive the text-based chatbots as more trustworthy if they are empathetic and customer-friendly during the interaction with the chatbots. Hence, we postulate the following hypotheses:

*H1.* Empathy of the text-based chatbot is positively related to consumers' trust toward the chatbot.

*H2.* Friendliness of the text-based chatbot is positively related to consumers' trust toward the chatbot.

*2.4 Moderators: task complexity and disclosure of the text-based chatbot*
*2.4.1 Task complexity.* The customer service in an e-commerce platform is supposed to help consumers to address their problems during the whole customer journey. The nature of customer service is to meet customers' needs across the stages of prepurchase, purchase and postpurchase (Lemon and Verhoef, 2016; Zhang *et al.*, 2016). In the prepurchase stage, consumers may need to search for information, recognize their needs and consider whether the product will meet their needs. During the interaction process, consumers may ask about the details, functions or discount strategies of the product itself. Thus, the task of the chatbot in this stage is to provide specific information about the products and assist the consumers with decision-making. In the purchase stage, consumers' behavior includes mainly choice, ordering and payment. During this process, consumers' interaction with the chatbot may be less than in the other two stages. In the postpurchase stage, consumers' behaviors mainly contain postpurchase engagement or service requests. According to previous studies, interaction between the store and consumer in this stage covers service recovery (Kelley and Davis, 1994), repurchase decision (Bolton, 1998), products returning (Wood, 2001) or other forms of engagement (van Doorn *et al.*, 2010). Tasks that the chatbot needs to address in this process may be more complex than the other two stages.

According to the task–technology fit theory (Goodhue and Thompson, 1995), the technology should meet the customer's needs or requirements for specific tasks (Blut *et al.*, 2021). Once the technology can meet the customer's specific needs, the interaction will be more satisfying. Therefore, in addition to the attributes of the chatbots in H1 and H2, more emphasis should be attached to the feature of the tasks that the chatbots address. Based on the task–technology fit theory, the attributes of the task can moderate the relationship between chatbots' attributes and customers' intention. Therefore, we focus on the effects of task complexity on customers' behavioral outcomes.

The effects of task complexity have been discussed in previous studies from several facets, such as task performance or team performance (Bjørn and Ngwenyama, 2009; Dayan and Di Benedetto, 2010). As aforementioned, tasks in different customer journey stages are different. Therefore, we applied task complexity in this research setting to better understand consumers' trust toward the chatbot in different contexts. When the problem is complex to address, consumers may turn to customer service for help. In this circumstance, they may pay more attention to the customer service providers' level of knowledge and professionalism. Thus, only giving rigid and standardized responses with friendliness and empathy cannot improve consumers' perceptions. In contrast, they may even doubt the chatbot's ability to address the problems. Consequently, we propose the following hypotheses:

> *H3a.* Task complexity moderates the relationship between empathy and consumers' trust toward the chatbot, such that the positive relationship is weaker when the task is complex.

> *H3b.* Task complexity moderates the relationship between friendliness and consumers' trust toward the chatbot, such that the positive relationship is weaker when the task is complex.

*2.4.2 Identity disclosure of the chatbot.* According to Davenport (2019), one way to increase trust in automation is to increase the disclosure of the system or its adoption as much as possible (Davenport, 2019). Consumers also have the right to be informed of the identity of the chatbot (Luo *et al.*, 2019). Disclosure in the e-commerce setting refers to the state when the text-based chatbots reveal their machine identity before interacting with consumers (Luo *et al.*, 2019). For example, in some online stores, the customer service will disclose its identity using the following sentences "Dear consumer, welcome to ABC store and I am the AI chatbot of this store" at the beginning of the interaction with them. Some stores also choose to state the AI identity beside the profile photo of the customer service staff. While a text-based chatbot can provide several business benefits for the company, some consumers may always feel uncomfortable talking with chatbots and perceive them as less empathetic than traditional customer service employees. While they are informed that they are interacting with a chatbot instead of a human, the positive effects of chatbots' empathy and friendliness will be increased due to the empathetic and friendly service. Accordingly, we propose the following hypotheses:

> *H4a.* Disclosure of the chatbot moderates the relationship between empathy and consumers' trust toward the chatbot, such that the positive relationship is stronger when the identity of the chatbot is disclosed.

*H4b.* Disclosure of the chatbot moderates the relationship between friendliness and consumers' trust toward the chatbot, such that the positive relationship is stronger when the identity of the chatbot is disclosed.

## 2.5 Response: avoidance/approach behavior toward the text-based chatbot

According to the SOR framework, consumers' response behavior consists of both averting behaviors and approach behaviors (Kim *et al.*, 2020). Extant studies mainly focus on consumers' approach behaviors or intention toward the automated bots. To provide a holistic view of consumers' response to the chatbots, we take both consumers' averting response (resistance) and approach response (reliance) to the text-based chatbot into consideration.

*2.5.1 AI reliance.* Individuals tend to rely more on a human rather than an algorithm for forecasting stock prices (Onkal *et al.*, 2009). Unlike humans, automated algorithms cannot provide reasons or explanations for their decision, thus increasing consumers' distrust. Although previous research has investigated consumers' future intention with the bots (Heerink *et al.*, 2010; de Kervenoael *et al.*, 2020), there has been little investigation and validation of practical interventions for the customer service provider to increase consumers' reliance on the automated machines instead of a human, especially in the e-commerce setting. Therefore, we tend to provide empirical evidence on consumers' reliance on the AI-based chatbot. As discussed in the literature on trust, cognitive trust involves individuals' confidence in others' performance and reliability. Therefore, we suggest that if consumers perceive the chatbot as trustworthy during the interaction process, they will be more likely to rely on the chatbot to address their problems or provide decisions. Thus, we hypothesize that:

*H5.* Consumers' trust toward the chatbot is positively related to their reliance on the chatbot.

*2.5.2 AI resistance.* Extant research on consumers' response to AI mainly focuses on their acceptance or future use intention of the AI-based bots (Ciechanowski *et al.*, 2019; Fridin and Belokopytov, 2014; Heerink *et al.*, 2010; Piçarra and Giger, 2018; Wirtz *et al.*, 2018). Although consumer pushback has been proven to be a major concern, only a few studies have investigated consumers' resistance to the AI-based bots (Longoni *et al.*, 2019). Findings in the healthcare context indicate that AI service providers are less likely to account for consumers' personal requests or unique characteristics. This can lead consumers to perceive the AI service provider as less trustworthy and they may exhibit resistant behavior toward them. In the e-commerce setting, when consumers find the text-based chatbots are not trustworthy or not able to meet their needs, they will avoid interacting with them and switch to the traditional human customer service (Longoni *et al.*, 2019). Therefore, the following hypothesis is proposed:

*H6.* Consumers' trust toward the chatbot is negatively related to their resistance to the chatbot.

The conceptual model is presented in Figure 1.

## 3. Research method

### 3.1 Sample and data collection

We measured the variables using existing scales adapted from previous research (see the Appendix). A pretest was conducted with 30 students in our research group to ensure the quality of the survey. Suggestions were also provided by these participants. After

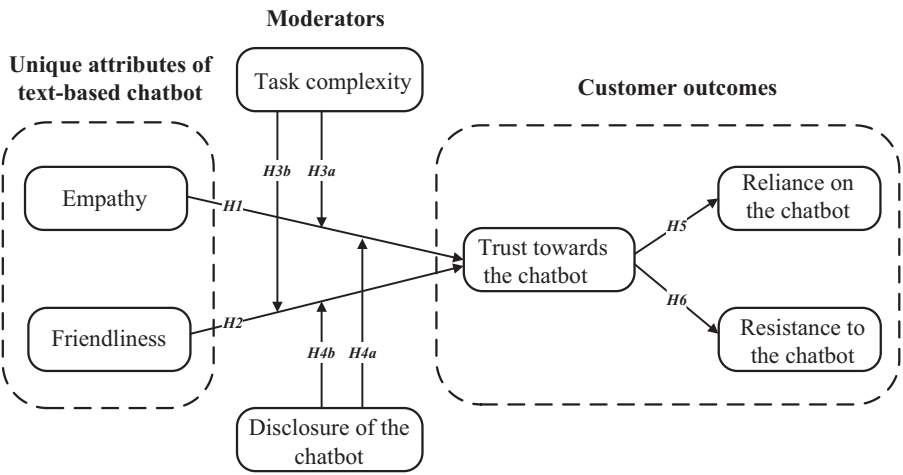

**Figure 1.**
A conceptual model of
customers' response to
the text-based chatbot

rearranging the survey, the formal version was then distributed mainly through a
professional survey website in China via a social network. Several representative chatbots in
the e-commerce setting were provided at the beginning of the survey. Those without any
chatbot experience during the purchasing process on e-commerce sites were filtered out. The
participants were told that their information was anonymous. Finally, a total of 441
respondents completed the survey including the pretest. After discarding incomplete and
rushed responses, there were a total of 299 effective responses, which were adequate for this
research. Our target participants were consumers who had experienced the text-based
chatbot in an e-commerce setting before.

Table 1 presents the demographics of these participants. As is shown in the table, the
participants consist of 28.8% males and 71.2% females. More than 82.9% of them are below
35 years old. The age range of the participants is consistent with this research setting because
young people nowadays interact with text-based chatbots more frequently in the online
purchasing process. Additionally, most of the participants are shown to frequently interact
with the text-based chatbots in the purchasing process. Therefore, the participants are
suitable for our research setting.

| Constructs and measurement items | Category | Frequency ($N = 299$) | % |
|---|---|---|---|
| Gender | Males | 86 | 28.8 |
| | Females | 213 | 71.2 |
| Age | <25 | 147 | 49.1 |
| | 26–35 | 101 | 33.8 |
| | 36–45 | 28 | 9.4 |
| | >45 | 23 | 7.7 |
| Experience with the chatbot in e-commerce | Seldom | 38 | 12.7 |
| | Less | 95 | 31.8 |
| | General | 106 | 35.5 |
| | More | 50 | 16.7 |
| | A lot | 10 | 3.3 |

**Table 1.**
Demographics of
participants ($N = 299$)

## 3.2 Measurements

The data was collected in China, so we conducted the forward–backward translation method to ensure the validity of translation (Lin *et al.*, 2019). We used multi-item measures with a seven-point Likert-style scale (1 = strongly disagree, 7 = strongly agree). The survey included seven reflective constructs, including the unique attributes of text-based chatbots (e.g. empathy, friendliness), moderators (e.g. task complexity, disclosure of the text-based chatbot), organism of the consumers' perception (e.g. trust toward the text-based chatbot) and responses of the consumers (e.g. resistance and reliance on the text-based chatbot). Control variables such as age, gender and consumers' experience with a text-based chatbot when purchasing online were also considered. Specifically, the *empathy* measure was developed following the guidance of previous research (Cheng *et al.*, 2018; Czaplewski *et al.*, 2002; de Kervenoael *et al.*, 2020). The *friendliness* measure was measured according to Tsai and Huang (2002). The *task complexity* measure was developed by adopting the measurements by Morgeson and Humphrey (2006). The *disclosure* measure was evaluated according to the information disclosure measure in extant research (Jensen and Yetgin, 2017; Luo *et al.*, 2019; Starzyk *et al.*, 2006). The *trust* measure was evaluated according to prior research on trust (Cyr *et al.*, 2009; Everard and Galletta, 2005). The *resistance* measure was developed following previous research on the resistance to information technology implementation (Kim and Kankanhalli, 2009; Lapointe and Rivard, 2005). The *reliance* measure was developed according to previous self-reliance measures (Edwards *et al.*, 2003).

## 3.3 Data analysis

We conducted the data analysis with SmartPLS 3.2.8 and SPSS 22.0. Specifically, SmartPLS was used to analyze the reliability, validity and common method variance (CMV) (Shiau *et al.*, 2019), while SPSS was used to test the hypotheses. The data analysis process was conducted in three stages: reliability and validity analysis, CMV and hypotheses testing.

*3.3.1 Reliability and validity analysis.* We first performed several statistical tests to ensure the reliability and validity of all the measures. As is shown in Table 2, we employed the value of Cronbach's $\alpha$ to test the internal consistency reliability. The Cronbach's $\alpha$ of all the measures exceeds the recommended threshold, indicating acceptable reliability. Moreover, the variance inflation factor (VIF) values of the subconstructs are all below the recommended threshold of 10, suggesting acceptable reliability (Hua *et al.*, 2020). The results of VIF also suggest that multicollinearity is not a significant concern in this research (Thatcher *et al.*, 2018). Validity of the construct measures was tested with two procedures. Firstly, each construct's average variance extracted (AVE) is above 0.5 and most of the constructs' factor loadings are above 0.7, suggesting satisfactory convergent validity. Discriminant validity was measured with the Fornell–Larcker criterion and cross-loading evaluation. As is shown in Table 3, the square root of AVE for each construct is greater than their correlations with other constructs (Liang *et al.*, 2019). As is shown in Table 4, each indicator's factor loading is greater than its cross-loadings with other constructs, suggesting satisfactory discriminant validity.

*3.3.2 Common method variance.* To address the concerns of CMV due to the self-reported surveys, we conducted several procedural and statistical remedies. In terms of procedural remedies, we tried to assure the respondents that they can answer these questions as honestly as possible to reduce the evaluation apprehension and protect their anonymity (Cheng *et al.*, 2020). In addition, we tried to improve the scale items by avoiding vague concepts and keeping the questions simple and specific (Podsakoff *et al.*, 2003). In terms of statistical remedies, we adopted the two most popular tests for the CMV, including Harman's one-factor test (Cram *et al.*, 2019; Saldanha *et al.*, 2017) and a marker variable technique (Griffith and Lusch, 2007; Hua *et al.*, 2020; Krishnan *et al.*, 2006; Lindell and Brandt, 2000) Specifically,

| Constructs | Items | VIF (<10) | Factor loadings (>0.7) | Cronbach's $\alpha$ (>0.7) | Average variance extracted (AVE) (>0.5) |
|---|---|---|---|---|---|
| Empathy (EMP) | EMP1 | 2.191 | 0.885*** | 0.863 | 0.785 |
| | EMP2 | 2.471 | 0.902*** | | |
| | EMP3 | 2.081 | 0.871*** | | |
| Friendliness (FRI) | FRI1 | 3.469 | 0.934*** | 0.926 | 0.870 |
| | FRI2 | 4.014 | 0.937*** | | |
| | FRI3 | 3.358 | 0.927*** | | |
| Task complexity (COM) | COM1 | 2.515 | 0.893*** | 0.881 | 0.807 |
| | COM2 | 3.511 | 0.918*** | | |
| | COM3 | 2.272 | 0.883*** | | |
| Disclosure of the chatbot (DIS) | DIS1 | 1.454 | 0.710*** | 0.717 | 0.733 |
| | DIS2 | 1.454 | 0.981*** | | |
| Trust toward the chatbot (TRU) | TRU1 | 1.337 | 0.671*** | 0.782 | 0.697 |
| | TRU2 | 2.415 | 0.919*** | | |
| | TRU3 | 2.161 | 0.893*** | | |
| Reliance on the chatbot (REL) | REL1 | 4.270 | 0.946*** | 0.951 | 0.910 |
| | REL2 | 5.675 | 0.957*** | | |
| | REL3 | 5.981 | 0.960*** | | |
| Resistance to the chatbot (RES) | RES1 | 1.767 | 0.792*** | 0.773 | 0.691 |
| | RES2 | 2.442 | 0.918*** | | |
| | RES3 | 1.574 | 0.776*** | | |

**Table 2.**
Psychometric properties of measures

**Note(s):** VIF – Variance Inflation Factor. The recommended threshold of each indicator is in the parentheses; ***$p$ < 0.001

Harman's single-factor test indicated no single factor. Then we conducted the "marker variable" test. Following Lindell and Brandt (2000), a marker variable is defined as a variable that is minimally correlated with the study variables. In this study, we chose age as a marker variable (Griffith and Lusch, 2007). The results suggest that the marker variable is not significantly related to all the variables in this model. Furthermore, the structural model paths of interest maintain statistical significance after including the marker variable. Therefore, CMV is not a serious problem in this study.

*3.3.3 Hypotheses testing.* We conducted the ordinary least squares regression to empirically estimate our proposed hypotheses. The results of the hypotheses tests are presented in Table 5 and Figure 2. As is shown in Table 5, Model 1 only contains the control variables. To answer the first research question (RQ1) (*What is the impact of the unique attributes of a text-based chatbot on consumer responses to the text-based chatbot in the*

| Construct | $M$ | SD | EMP | FRI | COM | DIS | TRU | REL | RES |
|---|---|---|---|---|---|---|---|---|---|
| EMP | 4.129 | 1.088 | *0.886* | | | | | | |
| FRI | 5.183 | 0.989 | 0.236 | *0.933* | | | | | |
| COM | 2.860 | 1.043 | −0.182 | −0.298 | *0.898* | | | | |
| DIS | 4.489 | 0.943 | 0.142 | 0.202 | −0.175 | *0.856* | | | |
| TRU | 4.266 | 1.000 | 0.753 | 0.262 | −0.227 | 0.194 | *0.835* | | |
| REL | 4.457 | 1.330 | 0.587 | 0.625 | 0.012 | 0.111 | 0.625 | *0.954* | |
| RES | 5.034 | 0.996 | −0.302 | 0.097 | −0.012 | 0.106 | −0.402 | −0.337 | *0.831* |

**Table 3.**
Mean, SD and discriminant validity evaluation based on Fornell–Larcker criterion

**Note(s):** Diagonal elements (italic) are the square root of AVE for each construct. Off-diagonal elements are the correlations between constructs; EMP: empathy, FRI: friendliness, COM: task complexity, DIS: disclosure of the chatbot, TRU: trust toward the chatbot, REL: reliance on the chatbot, RES: resistance to the chatbot

| Indicator | EMP | FRI | COM | DIS | TRU | REL | RES | |
|---|---|---|---|---|---|---|---|---|
| EMP1 | *0.885* | 0.216 | −0.220 | 0.165 | 0.679 | 0.522 | −0.299 | |
| EMP2 | *0.902* | 0.156 | −0.118 | 0.106 | 0.674 | 0.565 | −0.222 | |
| EMP3 | *0.871* | 0.256 | −0.145 | 0.106 | 0.648 | 0.473 | −0.281 | |
| FRI1 | 0.247 | *0.934* | −0.289 | 0.212 | 0.260 | 0.053 | 0.110 | |
| FRI2 | 0.197 | *0.937* | −0.304 | 0.192 | 0.224 | −0.046 | 0.078 | |
| FRI3 | 0.211 | *0.927* | −0.243 | 0.161 | 0.246 | 0.018 | 0.081 | |
| COM1 | −0.168 | −0.273 | *0.893* | −0.187 | −0.222 | 0.062 | −0.001 | |
| COM2 | −0.151 | −0.281 | *0.918* | −0.111 | −0.155 | 0.123 | −0.056 | |
| COM3 | −0.167 | −0.252 | *0.883* | −0.160 | −0.219 | −0.009 | 0.012 | |
| DIS1 | 0.009 | 0.169 | −0.132 | *0.710* | 0.059 | 0.034 | 0.116 | |
| DIS2 | 0.165 | 0.191 | −0.170 | *0.981* | 0.213 | 0.121 | 0.093 | |
| TRU1 | 0.481 | 0.426 | −0.286 | 0.326 | *0.671* | 0.248 | −0.120 | |
| TRU2 | 0.701 | 0.174 | −0.188 | 0.138 | *0.919* | 0.603 | −0.402 | |
| TRU3 | 0.677 | 0.156 | −0.147 | 0.102 | *0.893* | 0.628 | −0.412 | |
| REL1 | 0.549 | 0.050 | 0.043 | 0.069 | 0.612 | *0.946* | −0.334 | |
| REL2 | 0.560 | −0.024 | 0.066 | 0.117 | 0.591 | *0.957* | −0.337 | |
| REL3 | 0.572 | 0.006 | 0.056 | 0.133 | 0.586 | *0.960* | −0.294 | |
| RES1 | −0.264 | 0.286 | −0.163 | 0.110 | −0.317 | −0.421 | *0.792* | |
| RES2 | −0.283 | 0.045 | −0.006 | 0.106 | −0.367 | −0.323 | *0.918* | |
| RES3 | −0.203 | −0.077 | 0.134 | 0.048 | −0.315 | −0.097 | *0.776* | |

Consumers' response to text-based chatbots

**Note(s):** All italic values are significant with $p < 0.001$; EMP: empathy, FRI: friendliness, COM: task complexity, DIS: disclosure of the chatbot, TRU: trust toward the chatbot, REL: reliance on the chatbot, RES: resistance to the chatbot

**Table 4.**
Discriminant validity based on cross-loading evaluation

*e-commerce context?*), we contained the main effects in Model 2 and tested the impacts of empathy and friendliness on consumers' trust toward the text-based chatbot. The results of Model 2 indicate that empathy and friendliness are both positively related to consumers' trust (Model 2: $\beta = 0.652$, $p < 0.001$; $\beta = 0.093$, $p < 0.05$). Thus, hypotheses 1 and hypothesis 2 are supported.

To answer the second research question (RQ2) (*How does the task complexity and chatbot identity disclosure moderate the relationship between these attributes and consumer responses?*), we adopted a stepwise procedure and tested the moderation effects in H3a and H3b. Model 3 in Table 5 reports the moderation test results including the main effects and interaction effects. We mean-centered the construct scores and created the interaction terms of empathy, friendliness, task complexity and disclosure of the chatbot. After adding the interaction terms to the model, the results show that task complexity significantly moderates the relationship between friendliness and consumers' trust toward the chatbot (Model 3: $\beta = -0.073$, $p < 0.05$), thus supporting H3b. Using the same methods and procedures, we also tested for the moderation effects of disclosure of the chatbot on the relationship between chatbot attributes and consumers' trust. The results indicate that disclosure of the chatbot significantly moderates the relationship between empathy and consumers' trust (Model 3: $\beta = -0.154$, $p < 0.05$). In addition, disclosure of the chatbot also significantly moderates the relationship between friendliness and consumers' trust (Model 3: $\beta = 0.184$, $p < 0.01$), thus supporting H4b, while H4a is not supported.

To answer the third research question (RQ3) (*How does consumers' trust impact their responses to the text-based chatbot in e-commerce?*), we tested the relationship between trust and consumers' response (reliance and resistance) in Model 4, Model 5, Model 6 and Model 7. As for the relationship between consumers' trust and reliance on the chatbot, Model 4 only contains the control variables, while the results in Model 5 indicate the positive relationship between consumers' trust and reliance on the chatbot (Model 5: $\beta = 0.783$, $p < 0.001$), thus

**Table 5.**
Regression results

| Variables | | Trust toward the chatbot | | | Reliance on the chatbot | | Resistance to the chatbot | |
| --- | --- | --- | --- | --- | --- | --- | --- | --- |
| | Model 1 | Model 2 | Model 3 | Model 4 | Model 5 | Model 6 | Model 7 |
| *Main effects* | | | | | | | |
| Empathy | | 0.652*** (0.037) | 0.527*** (0.065) | | | | |
| Friendliness | | 0.093* (0.042) | 0.275*** (0.07) | | | | |
| Task complexity | | −0.079* (0.039) | −0.084* (0.04) | | | | |
| Disclosure of the chatbot | | 0.084* (0.041) | 0.008 (0.048) | | | | |
| Trust toward the chatbot | | | | | 0.783*** (0.061) | | −0.382*** (0.054) |
| *Interaction effects* | | | | | | | |
| Empathy * Task complexity | | | 0.024 (0.038) | | | | |
| Friendliness * Task complexity | | | −0.073* (0.032) | | | | |
| Empathy * Disclosure of the chatbot | | | −0.154* (0.067) | | | | |
| Friendliness * Disclosure of the chatbot | | | 0.184** (0.066) | | | | |
| *Control variables* | | | | | | | |
| Age | 0.097 (0.065) | 0.05 (0.043) | 0.051 (0.042) | 0.23** (0.085) | 0.154* (0.069) | 0.006 (0.065) | 0.043 (0.06) |
| Gender | −0.032 (0.130) | −0.034 (0.085) | −0.021 (0.084) | −0.097 (0.171) | −0.072 (0.137) | 0.067 (0.13) | 0.055 (0.12) |
| Experience with the chatbot | −0.021 (0.058) | −0.007 (0.038) | −0.005 (0.038) | 0.077 (0.076) | 0.093 (0.061) | −0.023 (0.058) | −0.031 (0.053) |
| Constant | 4.204*** (0.321) | 0.936* (0.384) | 0.776 (0.461) | 3.014*** (0.421) | −0.277 (0.425) | 4.971*** (0.32) | 6.58*** (0.372) |
| Observations | 299 | 299 | 299 | 299 | 299 | 299 | 299 |
| R-squared | 0.009 | 0.584 | 0.605 | 0.03 | 0.376 | 0.009 | 0.137 |

**Note(s):** Standard errors are in the parentheses; $*p < 0.05$, $**p < 0.01$, $***p < 0.001$

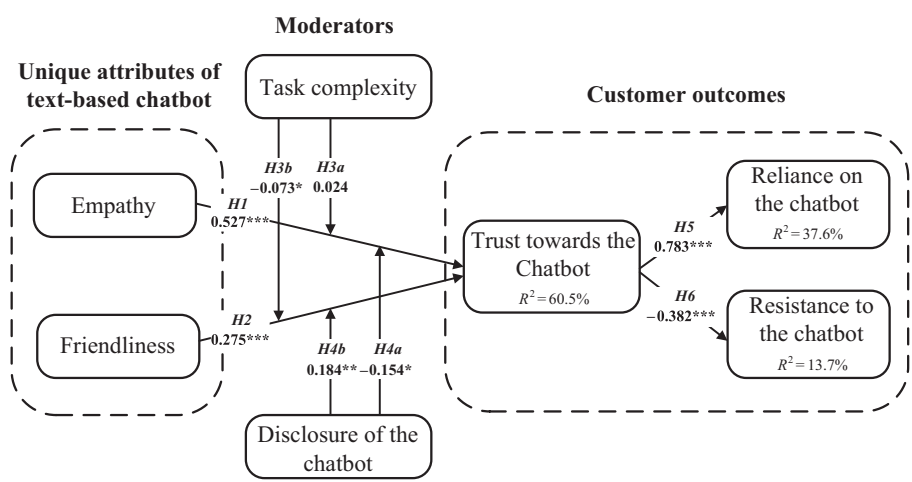

**Note(s)**: *p* < 0.05, **p* < 0.01, ***p* < 0.001

supporting H5. Similarly, for the relationship between consumers' trust and resistance to the chatbot, Model 6 contains the control variables of the research model. Finally, the results in Model 7 show that consumers' trust is negatively related to their resistance to the text-based chatbot (Model 7: $\beta = -0.382$, $p < 0.001$), thus supporting H6.

## 4. Discussions and conclusions

### 4.1 Summary of findings
Following Mehrabian and Russell's SOR framework (1974), the current study tests a model of consumers' trust and response to the text-based chatbot in the e-commerce setting, involving the moderating role of the different complexity of a task and the disclosure of the use of a chatbot. The specific findings of this study are summarized as follows:

Considering the first research question (RQ1), this study comes to the following findings. We mainly target the impacts of two specific attributes (empathy and friendliness) on consumers' responses in this study. Consumers' perceived empathy and friendliness of the text-based chatbot are both positively connected to their trust toward the chatbot. Specifically, the positive effect of empathy on trust is larger than that of friendliness. Our explanation for this finding is that, compared with a friendly manner, consumers prefer the chatbot to be able to appreciate their perspective and meet their specific needs. Therefore, empathy of the customer service provider is essential for successful interactions.

Considering the second research question (RQ2), this study provides empirical evidence on the boundary conditions of consumers' perception and response. We investigate the moderating effects of task complexity and text-based chatbot disclosure on the relationship between the chatbots' attributes and trust. The findings indicate that task complexity negatively moderates the relationship between friendliness and consumers' trust toward the chatbot, such that when the consumers' task is complex, the positive effect of friendliness on consumers' trust is weaker. Conversely, task complexity does not significantly impact the relationship between empathy and consumer' trust toward the chatbot. Our explanation for this finding is that, when the task is complex to address, consumers may be more concerned about the chatbots' professional level and ability to solve problems, instead of their attitudes or service manner. Considering the identity disclosure of the text-based chatbot, the findings indicate that the disclosure of the chatbot negatively moderates the relationship between empathy and consumers' trust toward the chatbot, while it positively moderates the relationship between friendliness and consumers' trust toward the chatbot.

Considering the third research question (RQ3), we take both consumers' approach response (reliance on the chatbot) and averting response (resistance to the chatbot) into consideration. Findings indicate that consumers' trust toward the chatbot is positively related to their reliance on the chatbot and negatively related to consumers' resistance to the chatbot. More specifically, the positive effect between consumers' trust and reliance is greater than the negative effect between consumers' trust and resistance. Thus, the effects of consumers' trust on their approach response are larger than that on their averting response.

### 4.2 Theoretical contributions

This study makes several theoretical contributions. First, this study contributes to the SOR framework by extending it to the automated chatbot in the e-commerce context. Previously, the SOR framework has mainly targeted traditional contexts (Kamboj *et al.*, 2018; Kim *et al.*, 2020). This extension contributes to the SOR framework by investigating specific attributes of automated text-based chatbots and how such attributes form consumers' trust and response in the e-commerce setting. The theoretical model in this research also contributes to the information systems literature.

Second, this study also considers the distinct characteristics of customer service in e-commerce by incorporating task complexity (Bjørn and Ngwenyama, 2009; Dayan and Di Benedetto, 2010) and chatbot identity disclosure (Luo *et al.*, 2019) as the boundary conditions of consumers' trust in this context. Prior research on consumers' response to the automated chatbot mainly discussed several attributes of the automated bots, such as anthropomorphism (Castelo *et al.*, 2019; Desideri *et al.*, 2019; Yuan and Dennis, 2019), humanness (Wirtz *et al.*, 2018) and adaptivity (Heerink *et al.*, 2010), the boundary conditions of these effects are largely ignored. The findings of this research show that when the tasks are different and the disclosure level is different, consumers' responses will also be different. Therefore, identifying the boundary conditions can also provide possible ways to reconcile the mixed findings in relevant research, thus opening an avenue for future study.

Third, this study provides a deeper understanding of consumers' responses to the automated AI-based chatbot, involving both approach response and averting response. As a new emerging technology, AI-based chatbots are providing the company and consumer with an expanding variety of benefits, therefore the effective application of automated bots is a big concern, especially in shaping the future of e-commerce. The extant literature on consumers' response to the AI-based bots has examined consumers' future intention with the automated machine (Ciechanowski *et al.*, 2019; Fridin and Belokopytov, 2014; Heerink *et al.*, 2010; Piçarra and Giger, 2018; Wirtz *et al.*, 2018). Therefore, there is a call for a systematic understanding of consumers' responses to a chatbot. This research adds to the understanding of both consumers' reliance and resistance to a text-based chatbot. Boundary conditions of consumers' responses are also identified, thus filling a gap in previous research that only focused on consumers' positive responses to the automated bots.

### 4.3 Practical implications

Our study also provides several practical implications for both customer service providers and the technology developers. First, as the AI-based chatbot is quite ubiquitous in the e-commerce setting, this research contributes to the limited understanding of consumers' response to the chatbot in the business context. Although the automated bots can arouse users' interest during the interactions, the online store should also be aware of consumers' intention to resist. The findings of this study can arouse the attention of the online stores that want to utilize chatbots more effectively than the human alternative but also more effective than their competitors.

Second, the findings of this study provide suggestions for the technology developer that creates and customizes the chatbot. This study provides empirical evidence for the effects of task complexity and identity disclosure. Specifically, consumers' perceptions of the chatbot will be different when their tasks are different. Interestingly, when the task is complex, the positive effects of chatbots' friendliness on consumers' trust will be weaker. Therefore, we suggest that it is more important for the chatbot to increase their professional level and address consumers' problems instead of merely replying to the consumers with standard answers in a friendly manner. Moreover, the disclosure of the machine identity should also be considered. Consumers' perceived disclosure of the automated chatbot may have different effects on the relationship between empathy/friendliness and trust. Therefore, the technology developer that customizes the chatbot should balance its characteristics to cater to consumers' different needs. Given the current capabilities of AI, this will require additional time and resources to further train and customize the chatbot, but it is worth it.

### 4.4 Limitations and future research

There are also limitations in this research. First, apart from empathy and friendliness, there are also other attributes of the chatbot that will affect consumers' responses, such as the intelligence level of the chatbot and passing the "Turing test" (Turing and Haugeland, 1950; Castelo *et al.*, 2019; Desideri *et al.*, 2019; Heerink *et al.*, 2010; Wirtz *et al.*, 2018; Yuan and Dennis, 2019). More specific characteristics of the chatbots will be taken into consideration in future research. Second, as there is a distinction between humans and machines, especially in the customer service institution, future research should also extend the understanding of the difference between human and machine customer service providers. Third, this study only investigated the chatbot in the e-commerce setting. While chatbots developed for the other domains (e.g. healthcare, sharing economy) may behave differently when compared to chatbots developed for the e-commerce setting. Specific issues need to be detected in other domains in future studies. Moreover, the characteristics of the customers may also have an impact on their acceptance or perception of the chatbot during purchasing. Therefore, attributes of customers should be taken into consideration in future research.

**Appendix**

| Constructs and measurement items | References |
|---|---|
| *Empathy*<br>(1) Text-based chatbots in e-commerce usually understand the specific needs of the customers<br>(2) Text-based chatbots in e-commerce usually give customers individual attention<br>(3) Text-based chatbots in e-commerce are available whenever it's convenient for customers<br>(4) If a customer requires help, the text-based chatbots would do their best to help (dropped) | Cheng *et al.* (2018), Czaplewski *et al.* (2002), de Kervenoael *et al.* (2020) |
| *Friendliness*<br>(1) The text-based chatbot had a kind service during our interaction<br>(2) The text-based chatbot provides the service in a friendly manner<br>(3) The text-based chatbot treats me nicely | Tsai and Huang (2002) |
| *Task complexity*<br>(1) The task I communicate with the customer service are simple and uncomplicated<br>(2) The task I communicate with the customer service comprises relatively uncomplicated tasks<br>(3) The task I communicate with the customer service involves performing relatively simple tasks | Morgeson and Humphrey (2006) |

**Table A1.**
Construct measurements

(*continued*)

| Constructs and measurement items | References |
| --- | --- |
| *Disclosure of the chatbot* | Jensen and Yetgin (2017), Luo *et al.* (2019), Starzyk *et al.* (2006) |
| (1) The identity of the text-based chatbot is disclosed before conversation | |
| (2) The e-commerce platform hides the identity of text-based chatbot from me | |
| (3) I am informed with the identity of the text-based chatbot before conversation (dropped) | |
| *Trust toward the chatbot* | Cyr *et al.* (2009), Everard and Galletta (2005) |
| (1) The text-based chatbot is honest and truthful | |
| (2) The text-based chatbot is capable of addressing my issues | |
| (3) The text-based chatbot's behavior and response can meet my expectations | |
| (4) I trust the suggestions and decisions provided by the text-based chatbot (dropped) | |
| *Reliance on the chatbot* | Edwards *et al.* (2003) |
| (1) I depend on the text-based chatbot for decision making | |
| (2) I depend on the text-based chatbot for service in e-commerce | |
| (3) I depend on the text-based chatbot for suggestions in purchasing process | |
| *Resistance to the chatbot* | Kim and Kankanhalli (2009), Lapointe and Rivard (2005) |
| (1) I prefer to use the human customer service rather than use the text-based chatbot | |
| (2) I will try to avoid using the service provided by text-based chatbot in e-commerce | |
| (3) I do not agree with the suggestions provided by the text-based chatbot in e-commerce | |

**Note(s):** Items with (dropped) were removed in the data analysis

**Table A1.**

## About the authors

Xusen Cheng is a professor of Information Systems in the School of Information, Renmin University of China, Beijing, China. He has obtained his PhD degree in Manchester Business School in the University of Manchester in the UK. His research focuses on information system and management, e-commerce and sharing economy, AI and behavior. His research paper has been accepted/appeared in journals such as *MIS Quarterly, Journal of Management Information Systems, European Journal of Information Systems, Decision Sciences, Tourism Management, Information & Management, Information Processing and Management, Group Decision and Negotiation, Information Technology and People, British Journal of Educational Technology, Computers in Human Behavior, Electronic Commerce Research, Internet Research,* among others. He serves as senior editor for Information Technology and People, associate editor for Information Technology for Development, Editorial Board for Electronic Commerce Research and Editorial Review Board for Internet Research, guest editor for several journals such as British Journal of Educational Technology, Electronic Markets, Journal of Global Information Technology Management and serves as track cochair and minitrack cochair for many conferences. His papers have also been presented in the leading conferences such as *International Conference of Information System (ICIS), Hawaii International Conference of System Science (HICSS) and American Conference of Information Systems (AMCIS).*

Ying Bao is a PhD candidate in the School of Information Technology and Management in the University of International Business and Economics, Beijing, China. Her interests focus on user behaviors and trust in the sharing economy and corporate social responsibility. Her research papers have appeared in *Information Processing and Management, Internet Research, Group Decision and Negotiation.* Her papers have also been presented in the leading conferences such as *Hawaii*

*International Conference of System Science (HICSS)* and *American Conference on Information System (AMCIS)*.

Alex Zarifis is a research associate in the School of Business and Economics at the University of Loughborough, UK. His first degree was a BSc in Management with Information Systems from the University of Leeds, followed by an MSc in Business Information Technology and a PhD in Business Administration both from the University of Manchester. His research focuses on the user's perspective in the areas of e-commerce, trust, privacy, artificial intelligence, blockchain, fintech, leadership, online collaboration and online education. His research has featured in journals such as *Information Processing and Management, Computers in Human Behavior, Information Technology and People, International Journal of Electronic Business* and *Electronic Commerce Research*. He served as the guest editor of *British Journal of Education Technology* and *Electronic Markets*. His research has been presented at conferences such as the European Conference on Information Systems (ECIS) and Hawaii International Conference on System Sciences (HICSS).

Wankun Gong is a postgraduate student in College of Humanities and Law in Beijing University of Chemical Technology, China. His research interests include e-government, e-service and e-commerce, human behavior and management, public and personal behavior analysis, trust and technology-enabled innovation.

Jian Mou is an associate professor of School of Business in Pusan National University, South Korea. His research interests include e-commerce, human–computer interaction, trust and risk issues in e-service and the dynamic nature in systems use. His research has been published in the *Journal of the Association for Information Systems*, *Internet Research, Information and Management, International Journal of Information Management, Computers in Human Behavior, Electronic Commerce Research, Information Processing and Management, Information Development, Behaviour and Information Technology, Journal of Retailing and Consumer Services, Electronic Commerce Research and Applications, Industrial Management & Data Systems, Journal of Global Information Management, IT & People, International Journal of Human–Computer Interaction* and *Information Technology and Tourism*, as well as in the proceedings such as *ICIS*, *ECIS*, *PACIS* and *AMCIS*. He serves as senior editor for Information Technology and People, associate editor for *Internet Research, and Behaviour and Information Technology*, Editorial Board for *Electronic Commerce Research* and guest editor *for several journals such as Information Technology for Development, Electronic Markets and Journal of Global Information Technology Management*. Jian Mou is the corresponding author and can be contacted at: jian.mou@pusan.ac.kr